\documentclass[journal]{IEEEtran}
\usepackage{amsmath}
\usepackage{amssymb}
\usepackage{url}
\usepackage{graphicx}

\begin{document}

\title{Efficient LiDAR Odometry for Autonomous Driving}

\author{Xin~Zheng,
        Jianke~Zhu,~\IEEEmembership{Senior~Member,~IEEE}
\thanks{Xin~Zheng and Jianke~Zhu are with the College of Computer Science, Zhejiang University, Hangzhou, China, 310027. \protect\\ 
E-mail: \{xinzheng,jkzhu\}@zju.edu.cn.}
\thanks{Jianke Zhu is the Corresponding Author.}}


\maketitle

\begin{abstract}
LiDAR odometry plays an important role in self-localization and mapping for autonomous navigation, which is usually treated as a scan registration problem. Although having achieved promising performance on KITTI odometry benchmark, the conventional searching tree-based approach still has the difficulty in dealing with the large scale point cloud efficiently. The recent spherical range image-based method enjoys the merits of fast nearest neighor search by spherical mapping. However, it is not very effective to deal with the ground points nearly parallel to LiDAR beams. To address these issues, we propose a novel efficient LiDAR odometry approach by taking advantage of both non-ground spherical range image and bird's-eye-view map for ground points. Moreover, a range adaptive method is introduced to robustly estimate the local surface normal. Additionally, a very fast and memory-efficient model update scheme is proposed to fuse the points and their corresponding normals at different time-stamps. We have conducted extensive experiments on KITTI odometry benchmark, whose promising results demonstrate that our proposed approach is effective.   
\end{abstract}

\begin{IEEEkeywords}
LiDAR odometry, autonomous driving, normal estimation, scan registration
\end{IEEEkeywords}

%
\IEEEpeerreviewmaketitle

\section{Introduction}

\IEEEPARstart{A}{utonomous} driving has attracted a large amount of research efforts due to the increasing needs in industry, where odometry plays an important role in building the high-precision map for navigation. Moreover, odometry is essential to accurate self-localization for path planning and environment perception, which is the key to driving safety. An effective odometry method should be robust to various environments including urban, highway and country road~\cite{cadena2016past}. In contrast to the approaches using video cameras~\cite{mur2015orb,zhu2017image}, LiDAR odometry is able to deal with large lighting variations by taking advantage of its active sensor emitting the laser beams. In this paper, we focus our attention on LiDAR-based method, which is widely used in practice. 

Generally, LiDAR odometry can be treated as a registration problem between the current scan and the reference point cloud model, which is effectively solved by Iterative Closed Point (ICP) algorithm~\cite{besl1992method}. As a local optimization-based method, ICP requires good initialization for better convergence. More importantly, the accuracy of pose prediction is highly dependent on the quality of inlier correspondences built by nearest neighbor search, which is usually the computational bottleneck in LiDAR odometry. 

To this end, the data structure such as KD-Tree~\cite{muja2009fast} is commonly employed to find the matched point pairs, which has the space complexity of $O(n)$ and the time complexity of $O(\log n)$. Although having achieved promising performance on KITTI odometry benchmark~\cite{geiger2012we}, the searching tree-based approach like LOAM~\cite{zhang2014loam} still has the difficulty in dealing with the large scale point cloud very efficiently. To guarantee the realtime performance, it adopts an aggressive down-sampling strategy that discards lots of useful observed points. 

Alternatively, the 3D voxel grid~\cite{biber2003normal} can be employed to efficiently search for the correspondences with the time complexity of $O(1)$. Unfortunately, it requires the storage complexity of $O(n^{3})$, which can hardly be used for the large scale outdoor scenarios due to huge memory consumption. One remedy is to make use of multi-scale representation like Octomap~\cite{hornung2013octomap} to take the trade-off between the memory usage and computational efficiency, which obtains the results inferior to the conventional searching tree-based method~\cite{zhang2014loam}.

\begin{figure}[t]
    \centering
    \includegraphics[width=0.5\textwidth]{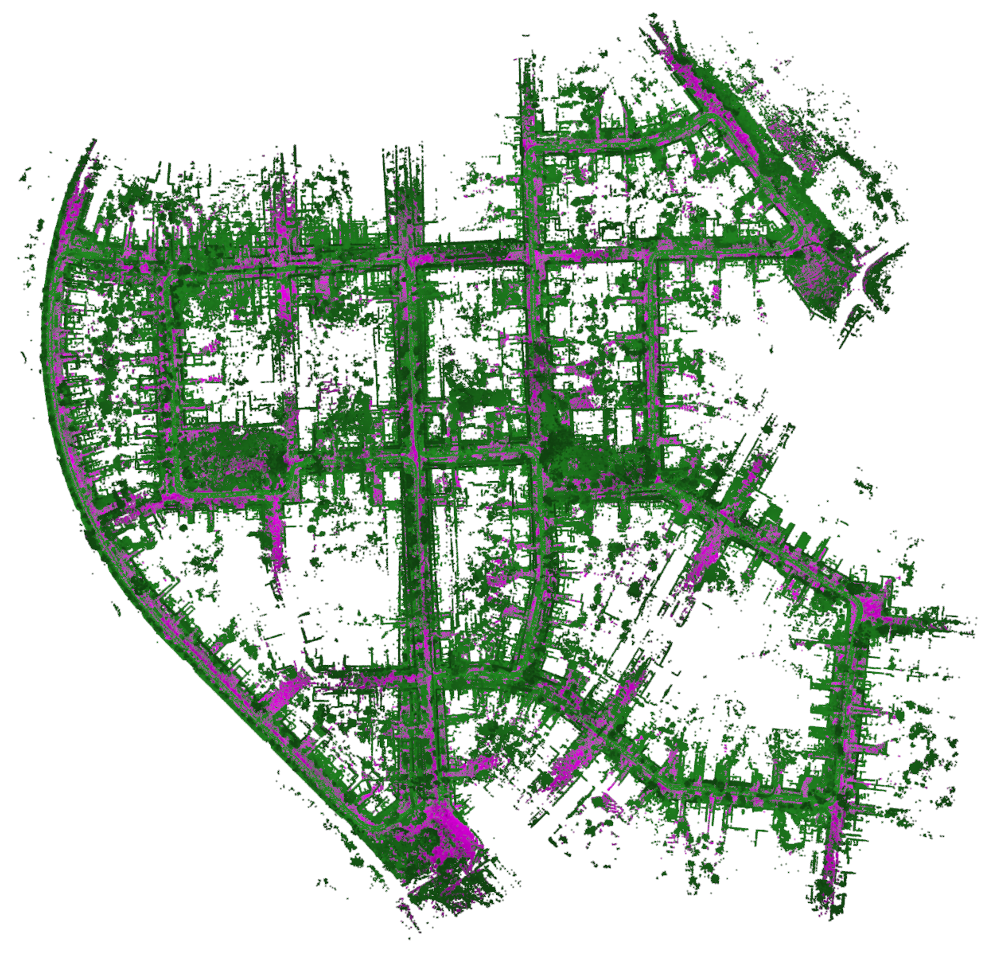}
    \caption{3D reconstruction results on KITTI dataset using our proposed efficient LiDAR odometry approach.}
    \label{fig:visual}
    \vspace{-0.1in}
\end{figure}

Most recently, Behley and Stachniss~\cite{behley2018efficient} directly map the scattered 3D point cloud onto a 2D spherical range image, where the correspondences can be very efficiently found within a small local patch. Although having the time complexity of $O(1)$ for nearest neighborhood search, their GPU implementation runs around 15 frames per second due to involving with the time-consuming surfel-based map update. They compute the normal map via the cross product of range image gradients, which is typically used in the dense depth map from RGBD sensor~\cite{whelan2015elasticfusion}.  Usually, the point density of LiDAR scans is far sparser than RGBD camera. They are relatively dense on the surface perpendicular to laser beams while being sparse in the region parallel to laser beams. Two adjacent points in spherical range image may be far away from each other, which are most likely belong to the different surfaces. Therefore, the cross product of range image gradients is not effective to compute normals, which inevitably leads to the inferior performance.

To address the above limitations, we propose a novel Efficient LiDAR Odometry (ELO) approach in this paper. In addition to the spherical range projection, we suggest to take advantage of bird's-eye-view (BEV) map that effectively retains the neighborhood relationship of ground surface points nearly parallel to laser beams. Thus, LiDAR odometry is formulated into a nonlinear least square minimization problem, where the non-ground cost and ground potential are adaptively fused according to their inlier ratios. Instead of using cross product, a range adaptive method is introduced to robustly estimate the local surface normals, which is computed by eigen-decomposition with outlier rejection. Additionally, a very fast and memory-efficient model update scheme is proposed to fuse the range image and their corresponding normal map at different time-stamps. Fig.~\ref{fig:visual} shows the reconstruction results of our method.

In summary, the main contributions of this paper are: 1) an efficient LiDAR odometry approach by taking advantage of both non-ground spherical range image and ground BEV map; 2) a robust range adaptive normal estimation method for LiDAR scan registration; 3) a very fast and memory efficient model update scheme that makes use of spherical range image and ground BEV map; 4) experiments on the KITTI odometry benchmark show that the proposed method not only achieves the promising results on LiDAR odometry but also runs over 169 frames per second.






\section{Related work}\label{sec:rel}

In general, the previous approaches to odometry can be roughly divided into two categories according to the different types of sensor. One group of research employs the video cameras~\cite{mur2015orb,zhu2017image}, which has been intensively studied for decade. It enjoys the merit of low cost and high resolution. Its main showstopper is that the video camera is sensitive to illumination changes and performs deficiently in the case of poor lighting conditions. Another group of methods are based on LiDAR~\cite{zhang2014loam,deschaud2018imls}, which directly recovers the 3D scattered points in the scene without resorting to the time-consuming triangulation or stereo rig. 

In this paper, we investigate the problem of LiDAR odometry that is typically treated as a scan registration problem in literature. Given the input point clouds perceived by LiDAR at two consecutive timestamps, the most popular solution for point cloud alignment is Iterative Closed Point (ICP) algorithm~\cite{besl1992method,brossard2020new,chebrolu2021adaptive}, which updates the transformation iteratively until convergence. Practically, the reliable correspondences between scans are essential to facilitate the effective ICP registration, where the nearest neighbor search is employed to obtain the matched points. To this end, various data structures are presented to efficiently find the closed points.  

Searching tree~\cite{muja2009fast} is widely used to retrieve the nearest neighbors with the space complexity of $O(n)$ and the time complexity of $O(\log n)$. The most successful LiDAR odometry approach LOAM~\cite{zhang2014loam} is based on KD-Tree, which has achieved low-drift with real-time performance in real-world applications. It selects the feature points by computing the roughness of a point on each scan line. Specifically, low roughness is denoted as planar feature while high roughness is denoted as edge feature. A variant of ICP algorithm combing both point-to-line~\cite{censi2008icp} and point-to-plane~\cite{low2004linear} metrics makes it reliable in different scenarios. To obtain real-time performance, LOAM predicts poses at 10Hz in frame-to-frame operation and adjusts drift at 1Hz in frame-to-model operation. Despite promising performance, simply dividing points by roughness cannot reflect the property of real geometry. Additionally, storing the local map points in KD-Tree~\cite{muja2009fast} is inconvenient for parallel computing. As in~\cite{demantke2011dimensionality}, Deschaud~\cite{deschaud2018imls} employs Principal Component Analysis (PCA) to select the salient planar features, which represents the map as implicit moving least squared surface. Although having achieved the encouraging results, it is very computational intensive, which cannot be deployed in realtime applications.

In~\cite{biber2003normal,ulacs20133d}, the 3D voxel grid-based approaches are able to find the nearest neighbors with the time complexity of $O(1)$, where each cell describes the geometry of local region. Similar to GICP~\cite{segal2009generalized}, it takes into account of point distribution in scan registration. The main drawnback of 3D grid map representation is that it consumes huge storage especially in the outdoor environment with the fine resolution.     


Behley and Stachniss~\cite{behley2018efficient} project 3D unordered LiDAR points onto the spherical range image, where the nearest neighbors can be very efficiently found within a small patch in 2D image space. Moreover, the surfel-based mapping~\cite{whelan2015elasticfusion} is introduced into the task of LiDAR odometry. The spherical range image of local map in the frame-to-model registration is rendered by an active surfel map. It is worthy of noting that mapping 3D scatter points onto a 2D image plane preserves the spatial neighborhood relationship that can be easily implemented by the efficient parallel computing technique.


Shan and Englot~\cite{shan2018lego} point out that nearly half observations from the LiDAR belong to ground regardless of the various scenarios in driving. Therefore, they present a lightweight ground-optimized LiDAR odometry method for the environment with the various terrain, which introduces the two-step optimization for pose prediction. This method imposes an extra ground constraint, which improves the reliability of LiDAR odometry in the structure-less circumstances such as highway. However, the road surfaces are usually not parallel to the heading direction of vehicles, which may not be effective to constrain the three aforementioned degrees of freedom with the ground features alone.



\section{Efficient LiDAR Odometry}~\label{sec:method}
\begin{figure*}[htbp]
    \centering
    \includegraphics[width=1.0\textwidth]{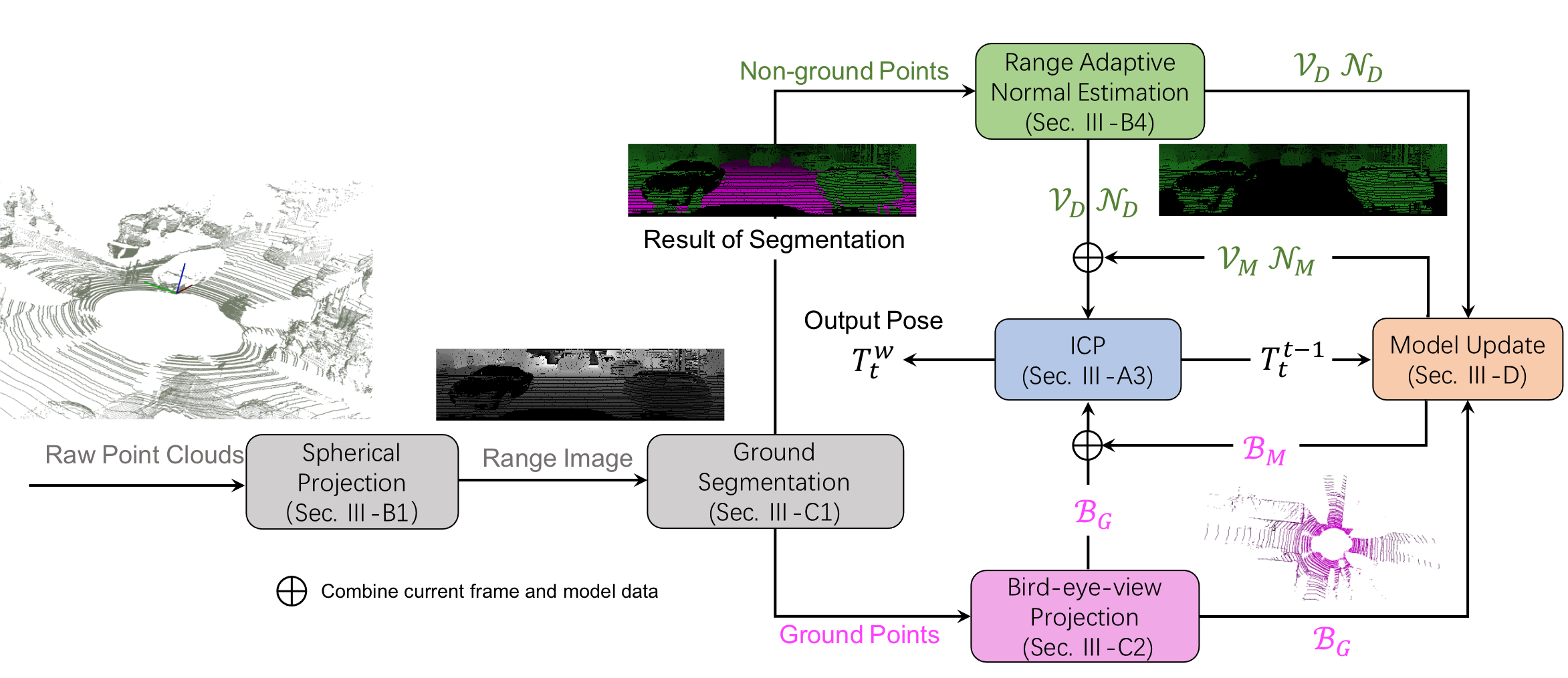}
    \caption{Overview of our proposed efficient LiDAR odometry approach.}
    \label{fig:pipeline}
\end{figure*}

In this section, we present our proposed approach to efficient LiDAR odometry for autonomous driving. Firstly, we formulate the problem into a non-linear least squares minimization problem with both non-ground and ground costs. Secondly, we present the non-ground cost using spherical range image and a range adaptive normal estimation method. Thirdly, we give the details on ground cost with 2D bird's-eye-view map. Finally, we suggest an efficient map update scheme for both non-ground range image and ground map.

\subsection{Efficient LiDAR Odometry for Autonomous Vehicles}
\subsubsection{Overview}
As in~\cite{zhang2014loam,shan2018lego,behley2018efficient}, LiDAR odometry is formulated as the frame-to-model registration problem, which aims at finding an accurate transformation between the consecutive scans.

As depicted in Fig.~\ref{fig:pipeline}, the raw 3D point cloud from LiDAR is firstly projected onto a spherical range image to facilitate the fast segmentation and non-ground cost (Section~\ref{sec:SRI}). Secondly, the ground points are segmented from the resulting spherical range image, which are further projected onto the 2D bird's-eye-view map to form the ground cost function (Section~\ref{sec:bev}). Thirdly, the normal map of spherical projection image is computed by range adaptive method (Section~\ref{sec:normal}), which is employed to estimate the pose increment by ICP. Finally, we update both the non-ground spherical range model and BEV ground map through the memory efficient update scheme  (Section~\ref{sec:map}).

\subsubsection{Proposed Fusion Approach}
To facilitate the effective registration, fast nearest neighbor search is the key to find the correspondences between the current scan and point could model, which is important to estimate the normal for calculating the point-to-plane error. As the computational bottleneck for LiDAR odoemtry, the searching-tree is commonly used. Inspired by the projective data association~\cite{behley2018efficient}, we directly map the scattered 3D points onto a 2D spherical range image to efficiently find the nearest neighbors, which is widely used in RGBD-SLAM ~\cite{izadi2011kinectfusion,whelan2015elasticfusion}. 

\begin{figure}[htbp]
    \centering
    \includegraphics[width=0.5\textwidth]{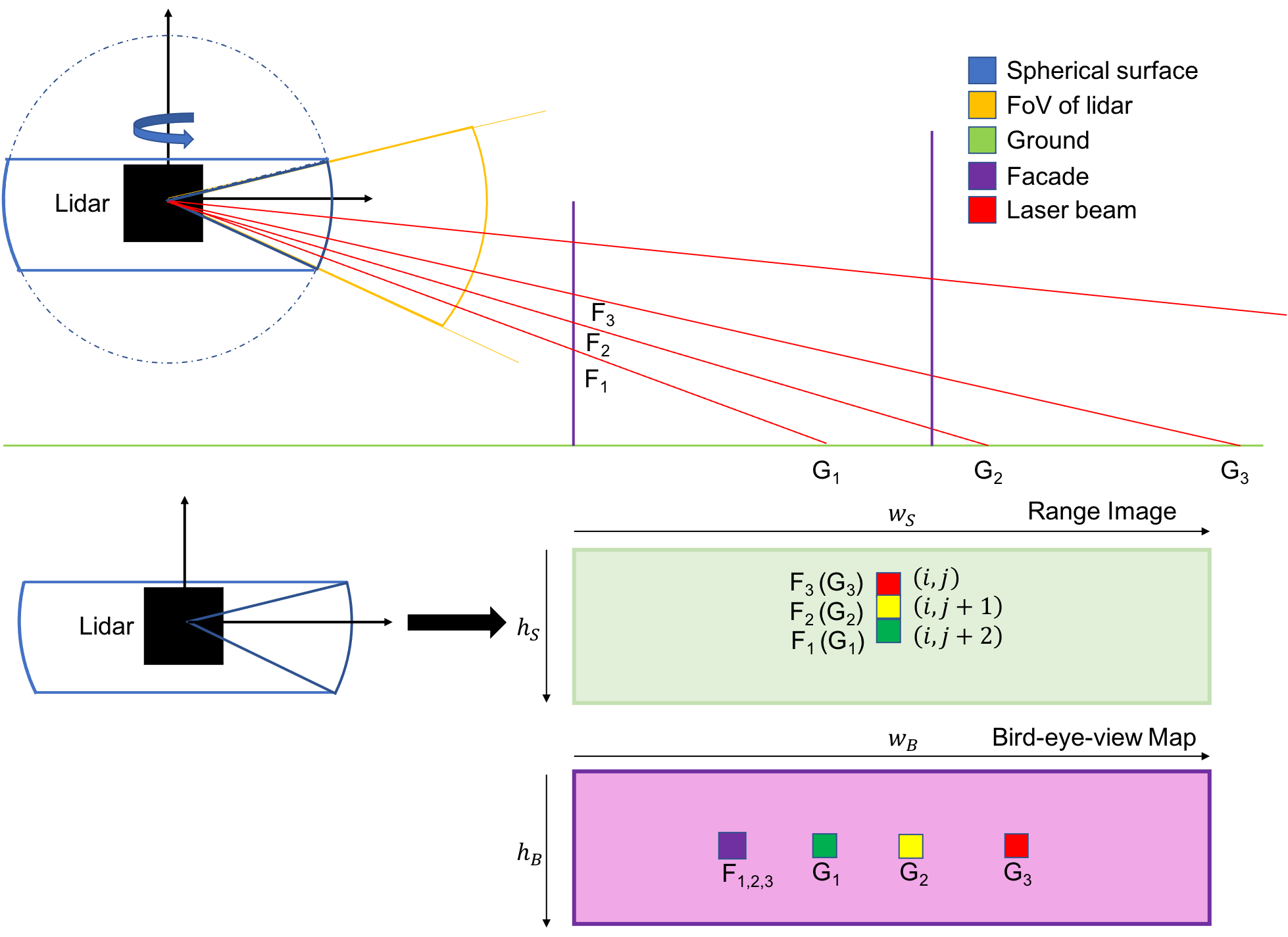}
    \caption{Example on the spherical projection for the facade points $F_{1},F_{2},F_{3}$ and ground points $G_{1},G_{2},G_{3}$.}
    \label{fig:projection}
\end{figure}

As discussed in~\cite{shan2018lego}, the orientation of LiDAR is slightly towards the ground in autonomous vehicles so that half of the scanned points belong to the road surface. Especially in the case of highway, there are few planar features extracted from the non-ground point cloud, which may lead to the drift for the pose estimation inevitably. Therefore, the non-ground features are insufficient for tracking while the ground points are good complement with the abundant planar surface features.

The distribution of perceived ground points is more sparser than non-ground ones. This incurs the issue that the adjacent pixels of ground points in spherical range image do not belong to the same local surface. We show an example in Fig.~\ref{fig:projection}. Given three adjacent laser beams, the intersections of facade and ground are $F_{1},F_{2},F_{3}$ and $G_{1},G_{2},G_{3}$, respectively. Their corresponding pixels in the spherical range image are in the same column. Points lying on the facade perpendicular to laser beams are the exact nearest neighbors on one surface. However, the distance between adjacent pixels can be very large when the reflection surface is parallel to laser beams. Moreover, spherical range image divides the space by angle, which means that plenty of ground points between $\overline{G_{1}G_{2}}$ or $\overline{G_{2}G_{3}}$ would project into same pixel especially long range ground points. This will lead to the incorrect correspondence assignments and large errors in estimating the normal of ground region by the conventional spherical range image. 






To deal with the above problem, we suggest to make use of bird's-eye-view~(BEV) map in order to taken into careful consideration of the abundant ground information, which greatly reserves the neighborhood relationship among ground points. As illustrated in Fig~\ref{fig:pipeline}, BEV map is a top-down projection capturing LiDAR points information. Let $z$-axis denote the altitude direction. Since the space is divided on $x-y$ plane, the neighborhood information of ground surface is kept in BEV map. Unfortunately, the surface perpendicular to laser beams like those facade points $F_{1},F_{2},F_{3}$ are condensed into one pixel in BEV map. To this end, we propose a fusion approach by taking advantage of both non-ground spherical range image and ground BEV map in this paper.

\subsubsection{Formulation} Assuming that the current scan is observed in LiDAR coordinate system $L_{t}$ and the model is represented in the coordinate system $L_{t-1}$, where the subscript $t$ is the timestamp. We aim at finding the transformation $T^{L_{t-1}}_{L_{t}}$ that efficiently aligns the current scan with previous model. To avoid gimbal lock affected by Euler angle representations, we parameterize the transformation by $T^{L_{t-1}}_{L_{t}}\in \mathrm{SE(3)}$, which is denoted as $T^{t-1}_{t}$ for simplicity. Specifically, we employ an adaptive weighting strategy to fuse the non-ground cost $E_{S}$ and ground cost $E_{G}$, which forms the following minimization problem:
\begin{equation}
    \min_{T^{t-1}_{t}} wE_{S}+(1-w)E_{G},
\end{equation}
where $w=w_{1}w_{2}$ is the product of two coefficients. $w_{1}=0.7$ is a user-defined weight to balance the number of non-ground and ground points, and $w_{2}$ is the inlier ratio between the non-ground points and ground ones. 


Obviously, the above problem is a nonlinear least squares minimization that can be efficiently solved through an iterative Gauss-Newton algorithm. Considering the large pose changes, we employ a constant acceleration scheme to initialize the transformation ${T^{t-1}_{t}}$. At each iteration, the pose update $\Delta T \in {\mathfrak{se}(3)}$ is estimated by $\Delta T =\left ( J^{T}J \right )^{-1}J^{T}\mathbf{e}$, where $\mathbf{e}$ denotes the residual vector. $J\in \mathbb{R}^{1\times 6}$ is weighted Jacobian matrix as below:
\begin{equation}
    J=wJ_{S}+(1-w)J_{G},
\end{equation}
where $J_{S}$ is the Jacobian of non-ground term, and $J_{G}$ is the Jacobian of ground term. The relative pose estimation $T^{t-1}_{t}$ is iteratively updated by $T^{t-1}_{t}=\exp(\Delta T)T^{t-1}_{t}$ until the convergence, where $\exp(\cdot):\mathfrak{se}(3) \mapsto SE(3)$ is the exponential map in Lie Algebra.

\subsection{Non-ground Cost $E_{S}$}~\label{sec:SRI} 
We give the details on non-ground cost $E_{S}$ that is computed by spherical range image. 
\subsubsection{Spherical Range Image~(SRI)}
The input data acquired from LiDAR is usually unordered 3D point clouds, which can be mapped into an organized 2D range image by spherical projection as in~\cite{izadi2011kinectfusion,whelan2015elasticfusion}. Therefore, the projection data association can be employed to very efficiently perform millions of nearest neighborhood search tasks in parallel. Given a scanned point $p=(x,y,z)^{T}$, the mapping function from the Cartesian to its corresponding spherical coordinate is defined as follows:
\begin{equation}
    \begin{bmatrix}
r\\ 
\theta 
\\ 
\phi 
\end{bmatrix}=\begin{bmatrix}
\sqrt{x^{2}+y^{2}+z^{2}}
\\ 
\arctan \left ( y/x \right )
\\ 
\arcsin\left ( z/\sqrt{x^{2}+y^{2}+z^{2}} \right )
\end{bmatrix},
\end{equation}
where $r$ denotes the range. $\theta$ is the azimuth, and $\phi$ represents the elevation component. They are constrained by $r \geq 0$, $-\pi < \theta \leqslant \pi$, and $-\pi/2 < \phi \leqslant \pi/2$. 


In this paper, the spherical range image is treated as a 2D search table $s(\theta,\phi)$, which stores the index of the Cartesian coordinates at azimuth $\theta$ and elevation $\phi$. The final image coordinates $(u,v)$ is converted by the spherical projection function $\Pi _{S}:\mathbb{R}^{3} \mapsto \mathbb{R}^{2}$,
\begin{equation}
    \begin{bmatrix}
u
\\ 
v
\end{bmatrix}=\Pi _{S}(x,y,z)=
\begin{bmatrix}
\frac{1}{2}\left ( 1-\frac{\theta}{\pi} \right )w_{S}
\\ 
\left [ 1-\left ( \phi+f_{up} \right )f^{-1} \right ]h_{S}
\end{bmatrix},
\end{equation}
where $f=f_{up}+f_{down}$ is the vertical field-of-view of the LiDAR sensor. $w_{S}=2048$ and $h_{S}=80$ are the width and height of spherical range image, respectively. 

\subsubsection{Non-ground Cost $E_{S}$ on Spherical Range Image}
We deal with non-ground points using spherical range image by taking advantage of efficient projective data association. As in~\cite{behley2018efficient}, the spherical range image is denoted as the vertex map $\mathcal{V}_{D}:\mathbb{R}^{2} \mapsto \mathbb{R}^{3}$, and its corresponding normal map is $\mathcal{N}_{D}$. Drifting usually occurs in the frame-to-frame registration due to error accumulation. To alleviate this issue, we adopt a frame-to-model method, where the model is represented by the vertex map $\mathcal{V}_{M}$ and normal map $\mathcal{N}_{M}$. Thus, we minimize the point-to-plane error to estimate the relative pose change as below:
\begin{equation}
    E_{S}(\mathcal{V}_{D},\mathcal{V}_{M},\mathcal{N}_{M})=\sum _{\mathbf{u}\in \mathcal{V}_{D}}\left [ \mathbf{n}^{T}_{u}\left (T^{t-1}_{t}\mathbf{u}-\mathbf{v}_{u}  \right ) \right ]^{2},
\end{equation}
where each vertex $\mathbf{u} \in \mathcal{V}_{D}$ is projectively associated to the model vertex $\mathbf{v}_{u}\in \mathcal{V}_{M}$ and its normal $\mathbf{n}_{u}\in \mathcal{N}_{M}$ via
\begin{equation}
    \mathbf{v}_{u}=\mathcal{V}_{M}(\Pi _{S}(T^{t-1}_{t}\mathbf{u})),
\end{equation}
\begin{equation}
  \mathbf{n}_{u}=\mathcal{N}_{M}(\Pi _{S}(T^{t-1}_{t}\mathbf{u})).
\end{equation}
Thus, the non-ground Jacobian $J_{S}$ can be computed as follows:
\begin{equation}\label{eqn:js}
    J_{S}=\mathbf{n}_{u}^{T}\begin{bmatrix}
 I& [\mathbf{v}_{u}]_{\times }
\end{bmatrix},
\end{equation}
where $[\mathbf{v}_{u}]_{\times}$ denotes the skew symmetric matrix of vector $\mathbf{v}_{u}$.

\subsubsection{Normal Estimation by Eigen Decomposition}
As shown in Eqn.~\ref{eqn:js}, the accuracy of Jacobian is heavily dependent on normal estimation, which is also essential to building the correspondences and computing the residual. Therefore, surface normal is the key to precisely predict the pose update. The conventional methods~\cite{whelan2015elasticfusion,behley2018efficient} simply compute the normal through the cross product of local image gradients that may not belong to the same local planar region. This incurs the large errors in normal estimation. To this end, we propose a novel robust range adaptive normal estimation with two outlier rejection criteria.


In general, a plane is defined by the equation $n_{x}x+n_{y}y+n_{z}z-d=0$, where $(x,y,z)^{T}$ lies on the plane and $(n_{x},n_{y},n_{z},d)$ are the corresponding plane parameters. Given a subset of 3D points $\mathbf{p}_{i}$, $i=1,2,\cdots ,k$ of the local surface, finding the optimal normal vector $\mathbf{n}=(n_{x},n_{y},n_{z})$ is a least square minimization on error $e$ as below:
\begin{equation}
    e=\sum_{i=1}^{k}\left ( \mathbf{p}_{i}^{T}\mathbf{n}-d \right )^{2} \textrm{ subject to } \left | \mathbf{n} \right |=1,
    \label{equ:normal}
\end{equation}
which has a closed-form solution by eigen-decomposition. The covariance matrix $\Sigma$ explains the geometric information of the given local surface, which is defined as follows:
\begin{equation}
    \Sigma=\frac{1}{k}\sum_{i=1}^{k}(\mathbf{p}_{i}-\bar{\mathbf{p}})(\mathbf{p}_{i}-\bar{\mathbf{p}})^{T}  ,
\bar{\mathbf{p}}=\frac{1}{k}\sum_{i=1}^{k}\mathbf{p}_{i},
\end{equation}
where $\bar{p}$ is the centroid of the point cloud subset. Using Eigen decomposition, $\Sigma\in R^{3\times 3}$ can be decomposed into three eigenvectors $\mathbf{v}_{1}$, $\mathbf{v}_{2}$ and $\mathbf{v}_{3}$, whose eigenvalues are in descending order $\lambda_{1}\geq \lambda_{2}\geq \lambda_{3}$. The least eigenvalue $\lambda_{3}$ indicates the variations along the surface normal $\mathbf{n}$ that is equal to its eigenvector $\mathbf{v}_{3}$. Moreover, we employ the surface curvature  $\sigma_{\mathbf{p}_{i}}$ at $\mathbf{p}_{i}$ to select the salience planar feature:
\begin{equation}
    \sigma_{\mathbf{p}_{i}}=\frac{\lambda_{3}}{\lambda_{1}+\lambda_{2}+\lambda_{3}},
\end{equation}

\subsubsection{Range Adaptive Normal Estimation}~\label{sec:normal}
With the spherical range image, we can visit a nearest neighbor with the time complexity of $O(1)$ for a small patch having the size of $(l_{x},l_{y})$, whose center location $\mathbf{p}$ is determined by its $(u,v)$ coordinates in range image. In the outdoor environment, the radius of LiDAR points have large variations within a scan, where the search window with the fixed size may not be effective. To account for the radius changes, we introduce a range adaptive searching window to select the nearest neighbors, in which the patch size is set according to the radius $r$, searching range threshold $\delta=0.3$m and range image resolution $(w_{S},h_{S})$:
\begin{equation}
    \binom{l_{x}}{l_{y}}=\binom{\max (\min (\frac{\delta}{r\pi}w_{S},l_x^{\max}),l_x^{\min})}{\max (\min (\frac{\delta}{rf}h_{S},l_y^{\max}),l_y^{\min})},
\end{equation}
where $l_{x}^{\max}=13$, $l_{y}^{\max}=7$, $l_{x}^{\min}=5$, and $l_{y}^{\min}=3$ denote the maximum and minimum searching range, respectively.

Due to the discontinuities at boundaries and multiple reflectance, the decomposition of covariance matrix is very sensitive to outliers. To reject the large outliers, we suggest two effective criteria. A point $\mathbf{p}$ is considered as an outlier if half points have the distance to the reference larger than the threshold $\sigma_{r}=0.5$m.

Another criterion is based on the point-to-plane distance $d_{p2p}=(q-p)^{T}\cdot \mathbf{n}_{p}$. A point is marked as outlier when its point-to-plane distance $d_{p2p}$ is large than the threshold $\sigma_{z} =0.5$, $d_{p2p}>\sigma_{z}$. 


\subsection{Ground Cost $E_{G}$}~\label{sec:bev}

We present the ground cost $E_{G}$ by taking advantage of bird's-eye-view map.
\subsubsection{Ground Segmentation}~\label{sec:groundSeg}
To efficiently obtain the ground segmentation, we employ a fast thresholding algorithm using the spherical range image. Since the height of LiDAR center with respect to road surface $h_{g}$ can be estimated from calibration, the ground point candidates need satisfy the condition $h_{p}-h_{g}> \delta_{h_{1}} $ and $h_{g}-h_{p}> \delta_{h_{2}} $. $h_{p}$ represents the height of candidate point. $\delta_{h_{1}}=0.5$m and $\delta_{h_{2}}=1$m are two thresholding values in height direction. 

Pixels in the column $v$ of spherical range image are regarded as lying in the same azimuth space. If two adjacent pixels are in ground, the angle change in height direction should be less than a small threshold $\delta_{\theta}=5^{\circ}$. A valid pixel at $(u,v)$ in range image indexes a 3D point $\mathbf{p}$ in Cartesian coordinate. We find the first valid points up $\mathbf{p}_{up}$ and down $\mathbf{p}_{down}$ whose coordinates in range image are $(u,v+i)$ and $(u,v-j)$, $i,j=1,2,3,\dots$. The angle change in two directions can be computed as below:
\begin{equation}
\theta_{up}=\arctan\frac{ \left \lfloor \mathbf{p}_{up} -\mathbf{p}\right \rfloor_{z}}{\left \lfloor \mathbf{p}_{up}-\mathbf{p} \right \rfloor_{xy}}, 
\theta_{down}=\arctan\frac{ \left \lfloor \mathbf{p}_{down}-\mathbf{p} \right \rfloor_{z}}{\left \lfloor \mathbf{p}_{down}-\mathbf{p} \right \rfloor_{xy}},
\end{equation}
where $\left \lfloor \cdot  \right \rfloor_{(\cdot )}$ denotes the normal length in the specific direction. The ground points are selected by the angle changes below the threshold $\theta_{up}<\delta_{\theta}$ and $\theta_{down}<\delta_{\theta}$. The remaining points are regarded as non-ground. 

\subsubsection{Bird's-eye-view (BEV) Map}
Similar to the spherical range image, bird's-eye-view projection also maps 3D Cartesian coordinate points into a 2D map. Specifically, each ground point $\mathbf{p}=(x,y,z)^{T}$ is converted into its corresponding coordinates $(u,v)$ in BEV map by the projection function $\Pi _{G}:\mathbb{R}^{3} \mapsto \mathbb{R}^{2}$ defined as follows:
\begin{equation}
    \begin{bmatrix}
u\\ 
v
\end{bmatrix}=\Pi _{G}(x,y,z) =\begin{bmatrix}
(x+w_{B})\cdot s_{x}
\\ 
(y+h_{B})\cdot s_{y}
\end{bmatrix},
\end{equation}
where $w_{B}=120$m and $h_{B}=60$m are the predefined scope in the two direction of BEV image. $s_x=0.1$m and $s_y=0.1$m represent the resolution of the BEV map in two directions.

In this paper, we suggest to make use of the 2D BEV map to effectively align the ground points of the current scan with the previous model map $\mathcal{B}_{M}$ through the projection data association. BEV vertex map $\mathcal{B}_{G}:\mathbb{R}^{2} \mapsto \mathbb{R}^{3}$ can be treated as a 2D search table for the fast nearest neighbor search, where each pixel contains the 3D ground points. Let $\mathbf{n}_{g}$ denote the surface normal for ground point, the cost function $E_{G}$ can be derived as below
\begin{equation}
    E_{G}(\mathcal{B}_{G},\mathcal{B}_{M})=\sum _{\mathbf{g}\in \mathcal{B}_{G}}[\mathbf{n}^{T}_{g}(T^{t-1}_{t}\mathbf{g}-\mathbf{v}_{g})]^{2},
\end{equation}
where each vertex $\mathbf{g}\in \mathcal{B}_{G}$ is projectively associated to a model vertex $\mathbf{v}_{g}\in \mathcal{B}_{M}$ via
\begin{equation}
    \mathbf{v}_{g}=\mathcal{B}_{M}(\Pi _{G}(T^{t-1}_{t}\mathbf{g})),
\end{equation}
The Jacobian $J_{G}$ for the ground cost $E_{G}$ can be computed by
\begin{equation}
    J_{G}=\mathbf{n}_{g}^{T}\begin{bmatrix}
 I& [\mathbf{v}_{g}]_{\times }
\end{bmatrix},
\end{equation}
where $[\mathbf{v}_{g}]_{\times}$ denotes the skew symmetric matrix of vector $\mathbf{v}_{g}$.

In contrast to the non-ground cost $E_{S}$, the surface normal $\mathbf{n}_{g}$ for ground point is not precomputed during the feature extraction stage. As the neighbors of ground points in a scan are usually far away from each other, the surface curvature of range image may not reflect the true geometry structure. Therefore, the normals of ground points are computed online while minimizing the ground cost $E_{G}(\mathcal{B}_{G},\mathcal{B}_{M})$. After projecting the current ground point $\mathbf{g}$ onto the previous local map $\mathbf{v}_{g}$, we retrieve the five closest points of $\mathbf{v}_{g}$ to calculate the surface normal in the BEV model map. By making use of the 2D BEV image, it is very efficient to find the candidate points in a predefined local patch on GPU. 

\subsection{Model Update Scheme\label{sec:map}}
Differently from the conventional data structure like voxel grid and searching trees, we suggest a very fast and memory efficient model update scheme that takes advantage of 2D spherical range image and ground BEV map. More importantly, it can be easily implemented in parallel.


In general, it can be clearly observed that the sampled data of laser beam is hardly to be the same point in the successive scans. Moreover, the LiDAR points are very sparse so that the surface features are inadequate within one scan. Therefore, the correspondences between the previous frame and current scan are usually not the ideal matches, where the errors will be accumulated. This may lead to the drift of vehicle pose estimation inevitably. To deal with this issue, we employ the frame-to-model scheme to fuse the consecutive scans. Once the pose change $T^{t-1}_{t}$ is estimated, we transform the local map of coordinate $L_{t-1}$ into the current coordinate $L_{t}$, and then integrate the point cloud in current frame to obtain a new local map. 

To reduce the influence of sensor noises, we record the timestamp of each point. Since the measurement noise of point cloud is highly related to its distance from the sensor, those long-range points have the high uncertainty. Once the vehicle travels forward, the early recorded points may introduce large errors in the pose estimation process. 
Therefore, we discard the old points whose timestamps exceed the time window $\tau_{w}=10$s, $t_{c}-t_{o}>\tau_{w}$. $t_{c}$ denotes the current timestamp, and $t_{o}$ is the observed timestamp in the model. 

We suggest two different updating strategies for the non-ground spherical range image and ground BEV map, respectively. As for non-ground point cloud, we maintain the model vertex map $\mathcal{V}_{M}$ and normal map $\mathcal{N}_{M}$. Given a valid pixel with the 3D position $\mathbf{p}$ in the previous spherical range image, we transform it by $T^{t-1}_{t}$, which is further projected into current image coordinate as pixel $(u,v)$. If the projected pixel is valid in the current vertex map $\mathcal{V}_{D}$, we compare the distance between the point $\mathbf{p}_{s}$ in current scan and the previous map point $\mathbf{p}_{m}$. The point close to LiDAR origin is chosen as $\min(\mathbf{p}_{s},\mathbf{p}_{m})$. Then, we update both vertex map $\mathcal{V}_{M}$ and normal map $\mathcal{N}_{M}$ with the selected point. If the location is not occupied, we simply cover it by the point in the previous map. For the ground points, we only maintain the BEV vertex model map $\mathcal{B}_{M}$ while the normal is computed with the five nearest points during ICP optimization. Similar to non-ground points, we just need to replace the spherical projection function $\Pi_{S}$ with the BEV projection function $\Pi_{G}$ in the updating process of ground model. Therefore, we only need keep three 2D maps for our proposed model update scheme, which is highly memory efficient.


\section{Experiment}~\label{sec:exp}

In this section, we present details of our experiments and discuss the results on LiDAR odometry. We test how effective the proposed approach is on the driving dataset. Additionally, we compare our presented method against the recent state-of-the-art approaches and evaluate on computational time.

\subsection{Experimental Testbed}
To investigate the efficacy of our proposed LiDAR odometry approach, we conduct the experiments on the KITTI odometry benchmark, where the 3D point scans are collected from the Velodyne HDL-64E S2. The whole dataset contains a wide variety of scenarios from urban city to highway traffic. There are 11 sequences (00-10) provided with GNSS-INS poses as the ground-truth for training, while another 11 sequences (11-21) without the ground truth for online evaluation on the leaderboard~\footnote{\url{http://www.cvlibs.net/datasets/kitti/eval_odometry.php}}. As in~\cite{deschaud2018imls}, a vertical angle of 0.195 degree is employed to correct the calibration errors in raw point clouds from the KITTI dataset. The average relative translation error $t_{rel}$ (\%) and rotation error $r_{rel}$ (deg/100m) are adopted as the performance metrics.


 
Our method is implemented by C++ with CUDA. All the experiments are performed on a laptop computer having an Intel Core i7-9750H CPU@2.60 GHz with 16GB RAM and an NVIDIA GeForce RTX 2060 GPU with 6GB RAM. Additionally, we evaluate the computational time of our proposed approach on NVIDIA Jetson AGX that is a popular embedded device for autonomous vehicles. 

\subsection{Evaluation on Different Projection Methods}

We firstly examine the performance on the different projection methods, including spherical range image, bird's-eye-view map, and our proposed fusion approach. To facilitate the fair comparison, all the methods employ our presented range adaptive approach to estimate the normals.

\begin{table}[htbp]
\addtolength{\tabcolsep}{-4pt}
    \scriptsize
    \caption{Performance Evaluation on Different Projection Methods}
    \label{tab:ground}
    \centering
    \begin{tabular}{cccccccccccc}
        \hline
        Approach &  00 & 01  & 02 & 03 & 04 & 05 & 06 & 07 & 08 & 09 & 10 \\
        \hline
        Spherical & 0.64& 0.73&0.59&0.93&0.45&0.48&0.33&0.56&0.87&0.56&1.42\\
        (all points)&0.28&0.17&0.22&0.26&0.42&0.27&0.14&0.35&0.26&0.17&0.61\\
        \hline
        Spherical & 0.71& 0.80&0.61&0.87&0.52&0.52&0.33&0.54&0.93&0.59&0.85\\
        (non-ground)&0.34&0.20&0.24&0.22&0.49&0.31&0.23&0.39&0.30&0.21&0.47\\
        \hline
        BEV map  & 0.76&0.66&43.02&64.46&0.27&5.07&12.60&118.90&1.03&127.73&165.00 \\
        (ground)& 0.33&0.17&15.08&43.34&0.20&10.90&0.40&42.52&0.34&17.84&40.97\\
        \hline
        Fusion  & \bf{0.54} & \bf{0.61} & \bf{0.54} & \bf{0.65} & \bf{0.32} & \bf{0.33} & \bf{0.30} & \bf{0.31} & \bf{0.79} & \bf{0.48} & \bf{0.59} \\
        (all points) & \bf{0.20} & \bf{0.13} & \bf{0.18} & \bf{0.27} & \bf{0.15} & \bf{0.17} & \bf{0.13} & \bf{0.16} & \bf{0.21} & \bf{0.14} & \bf{0.19}\\
        \hline 
    \end{tabular}
    \vspace{-0.1in}
\end{table}

        
\begin{table*}[t]
    \centering
    \scriptsize
        \caption{Results on KITTI Odometry Benchmark}
    \label{tab:benchmark}
\begin{tabular}{ccccccccccccccc}
\hline
Approach & 00 & 01 & 02 & 03 & 04 & 05 & 06 & 07 & 08 & 09 & 10 & Average & Online mean & Speed(s)\\
\hline
Frame-to-Frame&1.14&0.83&0.91&1.25&0.83&0.75&0.76&0.60&1.34&1.18&2.57&1.11&-&\bf{0.005}\\
&0.52&0.28&0.36&0.59&0.51&0.43&0.51&0.48&0.52&0.43&0.84&0.50&-&\\
\hline
Frame-to-Model& 0.54 & \bf{0.61} & 0.54 & \bf{0.65} & \bf{0.32} & 0.33 & \bf{0.30} & \bf{0.31} & \bf{0.79} & 0.48 & 0.59& \bf{0.50}& \bf{0.68} &0.006\\
& \bf{0.20} & \bf{0.13} & \bf{0.18} & \bf{0.27} & \bf{0.15} & \bf{0.17} & \bf{0.13} & \bf{0.16} & \bf{0.21} & \bf{0.14} & \bf{0.19} & \bf{0.18}& 0.21\\
\hline
Jetson & 0.55 &0.64 &0.56 & 0.70 &0.42 & 0.33 & 0.32 & 0.33 & 0.80& 0.47 &0.65&0.52&-&0.05\\
AGX &0.20 & 0.12 & 0.18 & 0.23 & 0.42& 0.17 & 0.14 & 0.17 & 0.23& 0.13& 0.21&0.20&-&\\
\hline
\hline
SuMa~\cite{behley2018efficient}&2.10&4.00&2.30&1.40&11.90&1.50&1.00&1.80&2.50&1.90&1.80& 2.93&-&- \\
Frame-to-Frame& 0.90&1.20&0.80&0.70&1.10&0.80&0.60&1.20&1.00&0.80&1.00& 0.92&-&\\
\hline
SuMa~\cite{behley2018efficient}    & 0.70 & 1.70 & 1.10 & 0.70 & 0.40 & 0.50 & 0.40 & 0.40 & 1.00 & 0.50 & 0.70 & 0.74&1.39&0.10\\
Frame-to-Model& 0.30 & 0.30 & 0.40 & 0.50 & 0.30 & 0.20 & 0.20 & 0.30 & 0.40 & 0.30 & 0.30 & 0.34& 0.34\\
\hline
SuMa++  & 0.64 & 1.60 & 1.00 & 0.67 & 0.37 & 0.40 & 0.46 & 0.34 & 1.10 & \bf{0.47} & 0.66 & 0.70  & 1.06 &0.10\\
 \cite{chen2019suma++} & 0.22 & 0.46 & 0.37 & 0.46 & 0.26 & 0.20 & 0.21 & 0.19 & 0.35 & 0.23 & 0.28 & 0.29  & 0.34 &\\
  \hline
  \hline
LOAM & 0.78 & 1.43 & 0.92 & 0.86 & 0.71 & 0.57 & 0.65 & 0.63 & 1.12 & 0.77 & 0.79 & 0.84 & 0.88 & 0.10\\
 \cite{zhang2014loam}& - & - & - & - & - & - & - & - & - & - & - & - & - & \\
 \hline
IMLS-SLAM & \bf{0.50} & 0.82 & \bf{0.53} &0.68 & 0.33 & \bf{0.32} & 0.33 & 0.33 & 0.80 & 0.55 & \bf{0.53} & 0.52 & 0.69 & 1.25  \\
 \cite{deschaud2018imls}& - & - & - &- & - & - & - & - &- &- & - & - & \bf{0.18} &   \\
 \hline
LeGO-LOAM   &2.17&13.4&2.17&2.34&1.27&1.28&1.06&1.12&1.99&1.97&2.21&2.49&-&-\\
\cite{shan2018lego}& 1.05&1.02&1.01&1.18&1.01&0.74&0.63&0.81&0.94&0.98&0.92&1.00&-&-\\
\hline
\end{tabular}
\vspace{-0.2in}
\end{table*}

As shown in Table~\ref{tab:ground}, it can be clearly seen that our proposed fusion approach obtains the best results on the KITTI training dataset, which indicates that the ground information is essential to the challenging scenario for LiDAR odometry, such as highway. Moreover, the method like~\cite{behley2018efficient} directly using spherical range image performs similar to the one with non-ground points alone. This implies that the non-ground points are effective enough to achieve the reasonable odometry results while the ground information is not full utilized in the spherical projection. Additionally, it can be observed that drifting occurs at several sequences with BEV map for the ground points. This is because the normal of surfaces perpendicular to laser beams cannot be accurately estimated using BEV map. It leads to the large rotation errors so that large drifts exist in the sequences with lots of quarter turns. Surprisingly, the results of BEV map outperforms the spherical range image method in the challenging case of highway on Sequence 01, where the ground region occupies large portion of scans. Therefore, the reliability of LiDAR odometry can be greatly improved by imposing the ground constraints. It can be concluded that our proposed fusion approach can fully exploit most of the local surfaces in the driving scenario.




\subsection{Evaluation on Normal Estimation}
We study the odometry results with the different normal estimation methods, including cross product of two point pairs, eigen-decomposition, and our proposed range adaptive approach. As shown in Table~\ref{tab:normal}, our presented range adaptive method consistently outperforms both cross product and plain eigen-decomposition at a very large margin. Moreover, the cross product method suffers from drifting issues in several sequences, since the adjacent points on the spherical range image have great potential lying on the different planes. Furthermore, the odometry results are improved a lot by the eigen-decomposition method.



\begin{table}[htbp]
    \centering
    \addtolength{\tabcolsep}{-2pt}
    \scriptsize
    \caption{Performance Evaluation on Normal Estimation}
    \label{tab:normal}
    \begin{tabular}{cccccccccccc}
        \hline
         &  00 &  01  & 02 & 03 & 04 & 05 & 06 & 07 & 08 & 09 & 10\\
        \hline
        Cross  &8.23&1.32&0.77&0.78&1.74&0.72&0.32&0.57&0.91&0.78&2.74\\
        Product& 3.66&0.31&0.32&0.33&0.41&0.34&0.21&0.35&0.35&0.27&1.17\\
        \hline
        Eigen& 0.85&0.98&0.97&1.62&1.33&0.62&0.59&0.80&1.42&1.03&1.39 \\
        Decomp.& 0.27&0.13&0.25&0.32&0.17&0.21&0.19&0.31&0.28&0.25&0.42\\
        \hline
        Range& \bf{0.54} & \bf{0.61} & \bf{0.54} & \bf{0.65} & \bf{0.32} & \bf{0.33} & \bf{0.30} & \bf{0.31} & \bf{0.79} & \bf{0.48} & \bf{0.59}\\
        Adaptive& \bf{0.20} & \bf{0.13} & \bf{0.18} & \bf{0.27} & \bf{0.15} & \bf{0.17} & \bf{0.13} & \bf{0.16} & \bf{0.21} & \bf{0.14} & \bf{0.19}\\
        \hline
        
        \vspace{-0.2in}
    \end{tabular}
\end{table}

\subsection{Comparison with State-of-the-art Methods}
To further examine the performance of LiDAR odometry, we compare our proposed method with the state-of-the-art approaches. For LOAM~\cite{zhang2014loam}, IMLS~\cite{deschaud2018imls}, SuMa~\cite{behley2018efficient}, SuMa++~\cite{chen2019suma++}, we directly include their results reported in the papers. Since there are no published results for Lego-LOAM~\cite{shan2018lego} on KITTI dataset, we conducted the experiments using their own implementation.


\begin{figure*}[htbp]
    \centering
    \includegraphics[width=1.0\textwidth]{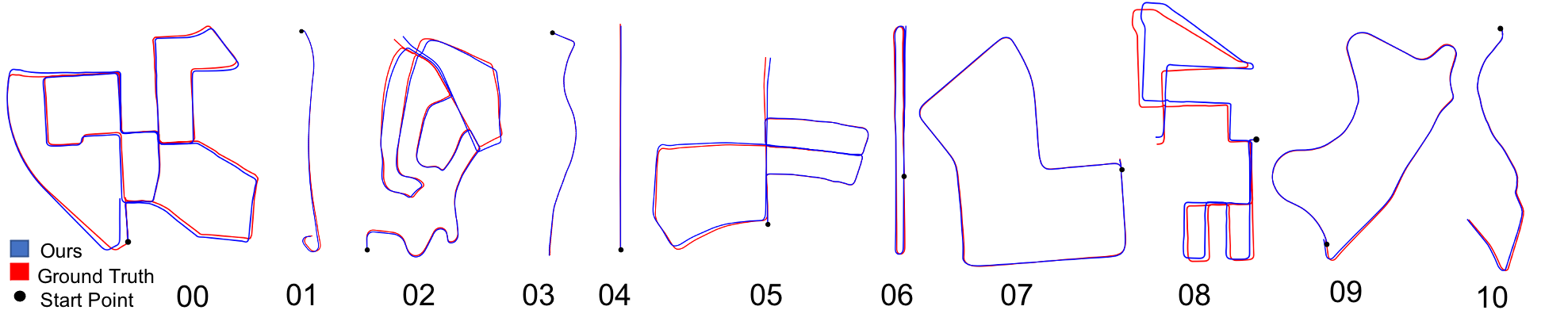}
    \caption{Camera path for our proposed method on KITTI odoemtry dataset.}
    \label{fig:trajectory}
    \vspace{-0.1in}
\end{figure*}

Table~\ref{tab:benchmark} shows the experimental results. It can be observed that our proposed approach achieves the best results with 0.50\% drift in translation and 0.0018 deg/m error in rotation on the KITTI training dataset. Fig.~\ref{fig:trajectory} plots the trajectory of our proposed method. Moreover, we obtain 0.68\% drift in translation and a 0.0021 deg/m error in rotation on the KITTI testing dataset, which outperforms IMLS with 0.69\% error in translation. Note that IMLS is the state-of-the-art LiDAR odometry method, which performs the best among the published results. Comparing to the spherical range image-based methods~\cite{behley2018efficient,chen2019suma++}, the proposed approach significantly improves the trajectory accuracy on KITTI benchmark by taking advantage of our presented projection fusion method and range adaptive normal estimation scheme. In contrast to the conventional two-step ground optimization method like LeGO-LOAM~\cite{shan2018lego}, our proposed bird's-eye-view map is very effective to capture the ground information.

\subsection{Evaluation on Computational Efficiency}


  

 The complexity of LiDAR odometry is mainly dominated by the nearest neighbor search and normal computation. To demonstrate the efficiency of our method, we evaluate the computational cost on different steps including spherical projection, feature extraction, ICP optimization, and model updating. Feature extraction is made of ground segmentation and non-ground normal estimation. The spherical projection is implemented on CPU with OpenMP, and the remaining steps are computed on GPU. Additionally, we implement our proposed approach on Nvidia Jetson AGX, which is commonly used in autonomous vehicles.

\begin{table}[htbp]
    \caption{Evaluation on Computational Time (ms)} \label{tab:runtime}
    \addtolength{\tabcolsep}{-4pt}
     \centering
     \scriptsize
\begin{tabular}{cccccc}
\hline
Approach & Spherical  & Feature  & ICP & Model & Total \\
& Projection& Extraction& Optimization & Updating & \\
\hline
PC frame2frame & 0.8 & 1.5&1.6 &0.3 & 4.2\\
\hline
PC frame2model&0.9&1.5&2.1&1.3 & 5.9\\
\hline
AGX frame2model&2.9&5.3&11.2&7.4 &26.8\\
\hline
\end{tabular}
\vspace{-0.1in}
\end{table}

As depicted in Table~\ref{tab:runtime}, the proposed approach runs about 169 frames per second on the commodity laptop computer while IMLS~\cite{deschaud2018imls} requires 1.25 second to process single scan. Moreover, the conventional spherical projection method~\cite{behley2018efficient} runs around 15Hz. This indicates that the presented scheme is not only very effective but also an order of magnitude faster than the conventional methods due to the efficient parallel implementation. It could be even faster with the less accurate frame-to-frame optimization, which runs at 238 frames per second. Finally, the presented method only requires 27 milliseconds on NVIDIA Jetson AGX, which could be potentially a key component for the autonomous vehicles.  



\section{Conclusion}~\label{sec:conc}
This paper proposed a novel efficient LiDAR odometry approach, which takes into account of both non-ground spherical range image and ground bird's-eye-view map. Moreover, the range adaptive method was introduced to robustly estimate the local surface normal. Additionally, we suggested an memory-efficient model update scheme to fuse the points and their corresponding normals at different time-stamps. We have conducted extensive evaluations on KITTI odometry benchmark, whose promising results demonstrated that our proposed approach not only outperforms the state-of-the-art LiDAR odometry methods but also runs over 169 frames per second on a commodity laptop computer.

Despite these encouraging results, some limitations and future work should be addressed. Currently, our method only takes considerations of LiDAR data. Besides, we have yet included the loop closure detection. For future work, we will address these issues by incorporating dense image alignment and backend optimization.


\ifCLASSOPTIONcaptionsoff
  \newpage
\fi



\bibliographystyle{IEEEtran}
\bibliography{mybib}

\begin{thebibliography}{10}
\providecommand{\url}[1]{#1}
\csname url@samestyle\endcsname
\providecommand{\newblock}{\relax}
\providecommand{\bibinfo}[2]{#2}
\providecommand{\BIBentrySTDinterwordspacing}{\spaceskip=0pt\relax}
\providecommand{\BIBentryALTinterwordstretchfactor}{4}
\providecommand{\BIBentryALTinterwordspacing}{\spaceskip=\fontdimen2\font plus
\BIBentryALTinterwordstretchfactor\fontdimen3\font minus
  \fontdimen4\font\relax}
\providecommand{\BIBforeignlanguage}[2]{{%
\expandafter\ifx\csname l@#1\endcsname\relax
\typeout{** WARNING: IEEEtran.bst: No hyphenation pattern has been}%
\typeout{** loaded for the language `#1'. Using the pattern for}%
\typeout{** the default language instead.}%
\else
\language=\csname l@#1\endcsname
\fi
#2}}
\providecommand{\BIBdecl}{\relax}
\BIBdecl

\bibitem{cadena2016past}
C.~Cadena, L.~Carlone, H.~Carrillo, Y.~Latif, D.~Scaramuzza, J.~Neira, I.~Reid,
  and J.~J. Leonard, ``Past, present, and future of simultaneous localization
  and mapping: Toward the robust-perception age,'' \emph{IEEE Transactions on
  robotics}, vol.~32, no.~6, pp. 1309--1332, 2016.

\bibitem{mur2015orb}
R.~Mur-Artal, J.~M.~M. Montiel, and J.~D. Tardos, ``Orb-slam: a versatile and
  accurate monocular slam system,'' \emph{IEEE transactions on robotics},
  vol.~31, no.~5, pp. 1147--1163, 2015.

\bibitem{zhu2017image}
J.~Zhu, ``Image gradient-based joint direct visual odometry for stereo
  camera.'' in \emph{IJCAI}, 2017, pp. 4558--4564.

\bibitem{besl1992method}
P.~J. Besl and N.~D. McKay, ``Method for registration of 3-d shapes,'' in
  \emph{Sensor fusion IV: control paradigms and data structures}, vol.
  1611.\hskip 1em plus 0.5em minus 0.4em\relax International Society for Optics
  and Photonics, 1992, pp. 586--606.

\bibitem{muja2009fast}
M.~Muja and D.~G. Lowe, ``Fast approximate nearest neighbors with automatic
  algorithm configuration.'' \emph{VISAPP (1)}, vol.~2, no. 331-340, p.~2,
  2009.

\bibitem{geiger2012we}
A.~Geiger, P.~Lenz, and R.~Urtasun, ``Are we ready for autonomous driving? the
  kitti vision benchmark suite,'' in \emph{2012 IEEE Conference on Computer
  Vision and Pattern Recognition}.\hskip 1em plus 0.5em minus 0.4em\relax IEEE,
  2012, pp. 3354--3361.

\bibitem{zhang2014loam}
J.~Zhang and S.~Singh, ``Loam: Lidar odometry and mapping in real-time.'' in
  \emph{Robotics: Science and Systems}, vol.~2, no.~9, 2014.

\bibitem{biber2003normal}
P.~Biber and W.~Stra{\ss}er, ``The normal distributions transform: A new
  approach to laser scan matching,'' in \emph{Proceedings 2003 IEEE/RSJ
  International Conference on Intelligent Robots and Systems (IROS 2003)(Cat.
  No. 03CH37453)}, vol.~3.\hskip 1em plus 0.5em minus 0.4em\relax IEEE, 2003,
  pp. 2743--2748.

\bibitem{hornung2013octomap}
A.~Hornung, K.~M. Wurm, M.~Bennewitz, C.~Stachniss, and W.~Burgard, ``Octomap:
  An efficient probabilistic 3d mapping framework based on octrees,''
  \emph{Autonomous robots}, vol.~34, no.~3, pp. 189--206, 2013.

\bibitem{behley2018efficient}
J.~Behley and C.~Stachniss, ``Efficient surfel-based slam using 3d laser range
  data in urban environments.'' in \emph{Robotics: Science and Systems}, vol.
  2018, 2018.

\bibitem{whelan2015elasticfusion}
T.~Whelan, S.~Leutenegger, R.~Salas-Moreno, B.~Glocker, and A.~Davison,
  ``Elasticfusion: Dense slam without a pose graph.''\hskip 1em plus 0.5em
  minus 0.4em\relax Robotics: Science and Systems, 2015.

\bibitem{deschaud2018imls}
J.-E. Deschaud, ``Imls-slam: scan-to-model matching based on 3d data,'' in
  \emph{2018 IEEE International Conference on Robotics and Automation
  (ICRA)}.\hskip 1em plus 0.5em minus 0.4em\relax IEEE, 2018, pp. 2480--2485.

\bibitem{brossard2020new}
M.~Brossard, S.~Bonnabel, and A.~Barrau, ``A new approach to 3d icp covariance
  estimation,'' \emph{IEEE Robotics and Automation Letters}, vol.~5, no.~2, pp.
  744--751, 2020.

\bibitem{chebrolu2021adaptive}
N.~Chebrolu, T.~L{\"a}be, O.~Vysotska, J.~Behley, and C.~Stachniss, ``Adaptive
  robust kernels for non-linear least squares problems,'' \emph{IEEE Robotics
  and Automation Letters}, vol.~6, no.~2, pp. 2240--2247, 2021.

\bibitem{censi2008icp}
A.~Censi, ``An icp variant using a point-to-line metric,'' in \emph{2008 IEEE
  International Conference on Robotics and Automation}.\hskip 1em plus 0.5em
  minus 0.4em\relax IEEE, 2008, pp. 19--25.

\bibitem{low2004linear}
K.-L. Low, ``Linear least-squares optimization for point-to-plane icp surface
  registration,'' \emph{Chapel Hill, University of North Carolina}, vol.~4,
  no.~10, pp. 1--3, 2004.

\bibitem{demantke2011dimensionality}
J.~Demantk{\'e}, C.~Mallet, N.~David, and B.~Vallet, ``Dimensionality based
  scale selection in 3d lidar point clouds,'' in \emph{Laserscanning}, 2011.

\bibitem{ulacs20133d}
C.~Ula{\c{s}} and H.~Temelta{\c{s}}, ``3d multi-layered normal distribution
  transform for fast and long range scan matching,'' \emph{Journal of
  Intelligent \& Robotic Systems}, vol.~71, no.~1, pp. 85--108, 2013.

\bibitem{segal2009generalized}
A.~Segal, D.~Haehnel, and S.~Thrun, ``Generalized-icp.'' in \emph{Robotics:
  science and systems}, vol.~2, no.~4.\hskip 1em plus 0.5em minus 0.4em\relax
  Seattle, WA, 2009, p. 435.

\bibitem{shan2018lego}
T.~Shan and B.~Englot, ``Lego-loam: Lightweight and ground-optimized lidar
  odometry and mapping on variable terrain,'' in \emph{2018 IEEE/RSJ
  International Conference on Intelligent Robots and Systems (IROS)}.\hskip 1em
  plus 0.5em minus 0.4em\relax IEEE, 2018, pp. 4758--4765.

\bibitem{izadi2011kinectfusion}
S.~Izadi, D.~Kim, O.~Hilliges, D.~Molyneaux, R.~Newcombe, P.~Kohli, J.~Shotton,
  S.~Hodges, D.~Freeman, A.~Davison \emph{et~al.}, ``Kinectfusion: real-time 3d
  reconstruction and interaction using a moving depth camera,'' in
  \emph{Proceedings of the 24th annual ACM symposium on User interface software
  and technology}, 2011, pp. 559--568.

\bibitem{chen2019suma++}
X.~Chen, A.~Milioto, E.~Palazzolo, P.~Giguere, J.~Behley, and C.~Stachniss,
  ``Suma++: Efficient lidar-based semantic slam,'' in \emph{2019 IEEE/RSJ
  International Conference on Intelligent Robots and Systems (IROS)}.\hskip 1em
  plus 0.5em minus 0.4em\relax IEEE, 2019, pp. 4530--4537.

\end{thebibliography}

%








\end{document}